# Computational analysis of pathological image enables interpretable prediction for microsatellite instability

*Jin Zhu, Wangwei Wu, Yuting Zhang, Shiyun Lin, Yukang Jiang, Ruixian Liu*, Xueqin Wang**

J. Zhu, W. Wang, Y. Zhang, Y. Jiang
Southern China Center for Statistical Science, School of Mathematics, Sun Yat-Sen University, 510275, Guangzhou, China.

S. Lin
Center for Statistical Science, Peking University, 100871, Beijing, China

Dr. R. Liu
Department of Clinical Laboratory, The Sixth Affiliated Hospital of Sun Yat-sen University, Guangzhou, Guangdong 510655, China
E-mail: liurx25@mail.sysu.edu.cn

Prof. X. Wang
Department of Statistics and Finance, School of Management, University of Science and Technology of China, Hefei, Anhui, 230026
E-mail: wangxq20@ustc.edu.cn

Microsatellite instability (MSI) is associated with several tumor types and its status has become increasingly vital in guiding patient treatment decisions. However, in clinical practice, distinguishing MSI from its counterpart is challenging since the diagnosis of MSI requires additional genetic or immunohistochemical tests. In this study, interpretable pathological image analysis strategies are established to help medical experts to automatically identify MSI. The strategies only require ubiquitous Haematoxylin and eosin-stained whole-slide images and can achieve decent performance in the three cohorts collected from The Cancer Genome Atlas. The strategies provide interpretability in two aspects. On the one hand, the image-level interpretability is achieved by generating localization heat maps of important regions based on the deep learning network; on the other hand, the feature-level interpretability is attained through feature importance and pathological feature interaction analysis. More interestingly, both from the image-level and feature-level interpretability, color features and texture characteristics are shown to contribute the most to the MSI predictions. Therefore, the classification models under the proposed strategies can not only serve as an efficient tool for predicting the MSI status of patients, but also provide more insights to pathologists with clinical understanding.

## 1. Introduction

Microsatellite instability (MSI) is the condition of genetic hypermutability that results from impaired DNA mismatch repair. Cells with abnormally functioning mismatch repair are unable to correct errors that occur during DNA replication and consequently accumulate errors. MSI has been frequently observed within several types of cancer, most commonly in colorectal, endometrial, and gastric adenocarcinomas.[1] The clinical significance of MSI has been well described in colorectal cancer (CC), as patients with MSI-high colorectal tumors have been shown to have improved prognosis compared with those with MSS (microsatellite stable) tumors.[2] In 2017, the U.S. Food and Drug Administration approved anti-programmed cell death-1 immunotherapy for mismatch repair deficiency/MSI-high refractory or metastatic solid tumors, making the evaluation of DNA mismatch repair deficiency an important clinical task. However, in clinical practice, not every patient is tested for MSI, because this requires additional polymerase chain reaction or immunohistochemical tests.[1] Thus, it is in high demand for a cheap, effective, and convenient classifier to assist experts in distinguishing MSI vs MSS.

Unfortunately, it is challenging to distinguish MSS from MSI based on medical expert's visual inspections from pathological images since the morphology of MSS is similar to that of MSI.[3] The recent technical development of high-throughput whole-slide scanners has enabled effective and fast digitalization of histological slides to generate whole-slide images (WSI). More importantly, the thriving of various machine learning (ML) methods in image processing, make this task accessible. Recent years, ML has been broadly deployed as a diagnostic tool in pathology.[4, 5] For example, Osamu Iizuka et al. built up convolutional neural networks (CNNs) and recurrent neural networks to classify WSI into adenocarcinoma, adenoma, and non-neoplastic.[6] The study by Yaniv Bar demonstrated the efficacy of the computational pathology framework in the non-medical image databases by training a model in chest pathology identification.[7] Notably, J.N. Kather et al. showed that deep residual network can predict MSI directly from H&E histology and reported the network achieved desirable performance in both gastric stomach adenocarcinoma (STAD) and CC.[8] These studies exhibit the great capacity of ML methods in medical research.

It is no doubt that the ML revolution has begun, but the deficiency of the 'interpretability' of ML deserves particular concern. Here, the concept of 'interpretability' means that clinical experts and researchers can understand the logic of decision or prediction outputted by ML methods.[9] Essentially, it urges ML systems follow a fundamental tenet of medical ethics, that is, the disclosure of basic yet meaningful details about medical treatment to patients.[10] Besides, interpretability helps clinician aware the decision provided by model would have potential fairness issue which is raised from the sampling bias in training models.[11] In addition, for both scientific robustness and medical safety reasons, interpretability allows researchers to know to what extent the predictions can be altered by small systematic perturbations to the input data, which might be generated by measurement biases. Finally, clinical experts are accessible to potentially crucial domain-knowledge hidden in the interpretable ML models. [9] Unfortunately, to the best knowledge of us, most of the existing MSI diagnosis systems, especially deep-learning based systems, are noninterpretable. Therefore, there is an urgent need to establish a new research paradigm for the application of interpretable ML system in medical pathology field. [12-16]

In this study, we used Haematoxylin and eosin (H&E)-stained WSI from The Cancer Genome Atlas (TCGA): 360 formalin-fixed paraffin-embedded (FFPE) samples of CC (TCGA-CC-DX),[17] 285 FFPE samples of STAD (TCGA-STAD) [18] and 385 snap-frozen samples of CC (TCGA-CC-KR). H&E stained images in these databases have already been tessellated into 108020 (TCGA-STAD), 139147 (TCGA-CC-KR), and 182403 (TCGA-CC-DX) color-normalized tiles, [8] and all of them only target region with tumor tissue. The aims of the study are: (i) to build an image-based ML method on MSI classification and post-process the fed image to a heat map so as to interpret the diagnosis of MSI at an image level; (ii) to design a fully transparent feature extraction pipeline and understand the pathological features' importance and interactions for predicting MSI by training a feature-based ML model.

## 2. Result

### 2.1. A Deep Learning Classifier and Image-Level Visual Interpretability

The deep learning (DL) model recruited to classify MSI is a famous end-to-end CNN, Resnet18.[19] To fit this model for different cancer subtypes, we trained three Resnet18 networks based on 70% tiles randomly sampled from three datasets. The remaining 30% tiles in each dataset were used for testing. At test time, a patient's slide was predicted to be MSI if at least half of the tiles were predicted to be MSI. The patient-level accuracy and area under the curve (AUC) was 0.81 in DX cohort, 0.80 in STAD cohort, 0.84 in KR cohort, respectively in the test set.

Based on the trained DL model, a novel method, Gradient-weighted Class Activation Mapping (Grad-CAM), can make the convolutional based model more transparent by generating localization maps of the important regions.[20] To unveil the hidden logic behind the DL and give a visual interpretability, we deployed Grad-CAM in our model to find out which part of the H&E image supports DL's classification. Two typical images for interpreting DL prediction logic are shown **(Figure 1.a)**. Under the pathologist's judgements, the highlighted region in Figure 1.a tends to be where tumor organism and immune number concentrate in and there exists a visible difference in the texture and color feature of both regions which inspired us to further collect detailed features in these directions to verify the validation of important region.

### 2.2. Transparent Pathological Image Analysis Workflow and Feature-Based Classification Model

The interesting information returned by Grad-CAM implies that important regions of the tumor organism might be encoded by certain features of the H&E stained images, which propelled us to study the classification capability of feature-based ML. To this end, we built an automatic and transparent workflow (Figure 2.a) with five main steps. We depict the first three steps in this part.

Firstly, we performed several image pretreatments for further study. White balance and brightness adjustments were performed to normalize the tone of the H&E stain before extracting features of all images. Meanwhile, color deconvolution[21] was applied to separate nucleus from the image for the next step. After the image pre-processing, we proceeded for visible pathological feature extraction. Motivated by the feedback from Grad-CAM, we focused on five H&E feature characteristics: global color feature extraction in RGB and HSV channels, local color feature extraction via gaussian mixture model (GMM) [22] with three components, the numbers of infiltrating immune cells and tumor cells, the grading of differentiation and the Haralick texture feature from tumor cells. Specifically, quantiles (25%, 50%, 70%) and moments (mean, variance, kurtosis, and skewness) in both RGB and HSV channels were recorded as global image features. In line with these, the moments of each cluster segmented by the GMM model were considered as local image features. In order to evaluate the grade of tumor differentiation, we used the circle Hough transform[23] in our workflow. In addition, we located the tumor cells and extracted their texture features via QuPath software[24]. Some typical feature extraction result is displayed in Figure 2.b. A total of 182 features were extracted from each image tile. The Wilcoxon rank sum test is applied on each feature and most of them are significantly different between MSI and MSS (Table S3).

We then applied Random Forest (RF),[25] one of the most popular ML algorithms, to all three databases next to classify MSI versus MSS on H&E stained histology slides. During training, 70% patients in every dataset were randomly selected and all of their tiles were used in training while the rest tiles were held out and used as test sets. In the test sets of each dataset, true MSS image tiles cohort has a median MSS score (the proportion of the prediction result judged to be MSS in each decision tree of the forest) is significantly different from those of MSI tiles (the P-value of the two-tailed test are 0.02, 0.0024, 0.002 in the three datasets), which means our models make sense in distinguishing MSI from MSS. Since one patient may have many different tiles, we obtain the MSI scores on patient-level through averaging the prediction on all its own tiles. AUCs for MSI detection are 0.7 (95% CI 0.65-0.74) in DX cohort, 0.74 (95% CI 0.65-0.79) in STAD cohort, and 0.78 (95% CI 0.7-0.82) in KR cohort (Figure 2.c). The decent performance of RF proves visible pathological features were proved to contribute to MSI prediction.

## 2.2. Feature-Level Visual Interpretability: Feature Importance and Interactions

One of the attractive advantages of the RF method is that it can evaluate the feature importance of the features. Therefore, we verify and quantify these features' power in distinguishing MSI from MSS by extracting information from a trained model. A common pattern can be discovered from the visualization of permutation-based feature importance[25, 26] in each dataset (Figure 3). From the figure, we can deduce that the Haralick Texture features play a dominant role. Besides, the texture features, which reflect the surface's average smoothness of the tumor cells in one tile, reveals that the tumor surface between MSI and MSS has significant differences. Color features also have significant contributions. In the global color feature, the higher order statistics (skew and kurtosis) contribute more than the first order statistics (mean and percentile), indicating that some useful information contributing to classification are hidden in high order features. Local color features generated from GMM also deserve our attention. As compared to global color features reflecting the global distribution, GMM features are regarded as the local ones since it can serve as an image segmentation and divide slices into different clusters so that we can obtain the information in each cluster. From Figure 2.b, the clusters generated from GMM highly correspond to tumor tissue or non-tumor tissue, which may make sense from the clinical perspective. Besides, the number of infiltrating immune cells also matters as expected while the differentiation grade gives the fewest contributions in each dataset. Except for the permutation method, we explored the contribution of these features with another feature importance criterion, mean minimal depth.[27] The importance ordering of features under this measure is highly consistent with the result from the permutation method (Figure S1), strengthening the common feature importance pattern we discovered.

It is widely accepted that feature interactions (i.e., the joint effect of features) is nonnegligible for the complex disease.[28] Our feature-based RF models also allow us to exploit the pairwise feature interactions in MSI classification, and thus, we can attain more clear understanding about the characteristics of MSI tiles as well as the mechanism of RF. Here, we use condition minimal depth[27] to quantitatively assess feature interaction and then demonstrate the foremost 30 pairwise interactions (Figure 4). The features with the most significant interaction effect with other features in each dataset are: the skew of S channel in DX, the variance of H channel in the second component of GMM in KR, and the optical density mean of the nucleus of tumor cells in Hematoxylin stain in STAD. The three features enhance the importance of the feature interacted with them, even the feature may not be significant before. To concretely comprehend how the paired features jointly help the prediction of our RF model, we plot the prediction values of typical feature interaction on a grid diagram (Figure S2). In DX, a higher value of the max caliper in tumor cells and a lower value of tumor cell count leads to a higher probability of MSS prediction. In the KR dataset, a higher value of immune cell number and a lower value of the 75th percentile of R channel leads to a higher probability of MSS prediction. In STAD, a lower value of the optical density range of tumor cells' nucleus in Hematoxylin stains and a higher value of Haralick correlation in Eosin stains leads to a higher probability of MSS prediction. In conclusion, we can say that the interaction effects are vital for MSI diagnosis and tend to occur more often between color features and texture features.

## 3. Conclusion and Discussions

To our knowledge, this is the first study to not only build up a classification model in distinguishing MSI from MSS but also provide a detailed interpretability analysis and unveil the hidden logic behind the model. In this study, we followed the framework of interpretability with two steps: first, built up a high-performance DL network with a visual explanation capacity as model-based interpretability; secondly, we further analyzed and confirmed features' power using a feature-based interpretable model. The datasets in our study come from TCGA and are comprised of three different cancer types which are DX, KR, and STAD, respectively. The images we used have already been tessellated into small tiles and gone through the color normalized process.

To achieve the target of building up a DL network with visual explanations, we trained residual learning CNNs and deployed Grad-CAM to the final convolutional layer of the network so as to produce the heatmap that reflects the highly-contributed region. The network can achieve a state-of-the-art performance. Notably, through its coarse localization map of the important regions in the image, it provided a preliminary insight to the pathologists about highly-contributed pathological features.

To further verify and quantify the related features' power in classification, we designed a pipeline to extract the features which may make sense in these regions from images, train and evaluate a random forest classifier to distinguish MSS from MSI and applied different measures to evaluate the contribution of each feature so that we can supply interpretations for our model. We extracted 182 features in total from images and these features capture color features locally and globally, the count of immune cell and tumor cell, differentiation degree of tumor tissue and Haralick texture features of tumor cells. Two different techniques, permutation and mean minimal depth, were adopted to study the contribution of features and a common importance pattern was discovered. Besides, the interaction between features based on trees' structure in the random forest was also explored. The analysis of the features' contribution in classification and the existing interaction between them can help to unveil the hidden logic in distinguish MSS from MSI. To the best of our knowledge, it is the first study shows the texture and color of the H&E image and their interactions are crucial for the diagnosis of MSI.

We believed our work can be deployed at medical center to facilitate MSI diagnosis at a low cost based on H&E stained histopathology slides. The localization map outputted by our DL models can help experts to narrow their focus on the specific region of the whole H&E slide, thereby contributing to a more accurate diagnosis with the prediction result of our model. Besides, the features' distribution under our interpretable model can provide experts with more insight into analyzing the slices of MSI and MSS from clinical perspectives. Further, considering the similar feature distribution pattern in three datasets we used, it is possible that after running the same pipeline on MSI H&E slides under different cancer types, we can discover a generalization pattern behind them. After training on a larger dataset, the accuracy of the identification and the interpretability could get improved, thereby contributing to accurate sample curation and treatment development of this aggressive cancer subtype.

Previous studies in investigating the pathologic predictors of microsatellite instability through feature extraction and logistics regression model suffered from the limited learning capability of the model and the small sample size, and thus could not achieve great performance.[29] Other works about CC study indicated the significance of the histological structures inspired our work in feature extraction.[30, 31] Another work on MSI classification paid attention to the enhancement of the prediction accuracy through establishing a DL network but did not give a detailed description of the mechanism behind the model.[8]

Moreover, we conjecture that the dominant-role features in RF models have a significant importance in the DL model. To verify our speculation, we eliminated the RGB mean differences between MSI and MSS groups and reevaluated the AUCs of our DL models. More precisely, we calculate the RGB mean value of all the tiles in both groups and regulate the RGB mean value of every tile into that population mean value. We found that the AUCs reduced 0.11, 0.12, and 0.14 in DX, KR, and STAD datasets, respectively, which verifies that RGB features also make contributions to the DL method.

One limitation of this study is that the cases in TCGA datasets may not be an unbiased collection from the real situation since pathologists may only upload the representative ones. Although our model performed well in these histopathology images, we should admit that their performance in the actual clinical settings requires a further research. Another limitation is that our study only focused on H&E stained images and we could not confirm whether the pattern in this study especially the color features' contribution works in other types of histopathology slices. The classifier model under other types i.e. immunochemical stained images remains to be explored and established.

In summary, we developed ML models with decent power in the prediction of MSI. Moreover, our models exhibit visual heatmap demonstrating high-contribution regions for MSI prediction in the H&E image, and we certified certain pathological features have non-trivial importance in MSI classification, which is not explicitly studied in the previous research. Therefore, our study not only facilitate MSI diagnosis based on H&E image but also shed a light on the understanding of MSI at both image level and features level. As a by-production of our study, a user-friendly and ongoing-upgraded web application (http://14.215.135.56:3838/DL/) was developed for world-wide clinical researchers.

## 4. Methods

*Histopathology Image Sources:* The whole-slide H&E stained histopathology images were obtained from TCGA, which included 3 cancer subtype datasets. Dataset DX consisted of 295 MSS patients and 65 MSI patients from FFPE samples of CC. Dataset KR contained 316 MSS patients and 72 MSI patients from snap-frozen samples of CC. Dataset STAD collected 225 MSS patients and 60 MSI patients of FFPE stomach adenocarcinoma.

All the images used in our models have already gone through tumor tissue detection and been tessellated into small tiles in J.N. Kather's work (https://zenodo.org/record/2530835 and https://doi.org/10.5281/zenodo.2532612). There are 108020 tiles in TCGA-STAD cohort, 139147 in TCGA-CC-KR, and 182403 in TCGA-CC-DX. Color normalization has already been performed on every tile using Macenko method, which converts all images to a reference color space.

*The Details of Grad-CAM:* Grad-CAM utilizes the gradient information abundant in the last convolutional layer of a CNN and generates a rough localization map of the important regions in the image. We apply rectified linear unit to the linear combination of maps so as to generate localization maps of the desired class. Grad-CAM visualization was implemented in Python 3.7 with TensorFlow-GPU 1.14.0 and Keras 2.3.0.

*The Details of Image Pretreatment:* We apply pretreatments to the tiles before feature extraction. First, in order to avoid the influence of color cast, the natural appearance tone of the object is altered in the formation of images when exposed in a lightning condition of different color temperature, white balance is performed on our cohorts. Since every tile has area without cell organization, i.e. without H&E stained, we could view that part as the neutral reference in adjustment. Besides the color cast, overexposure and underexposure also may result in the distortion of our features. Still, taking the unstained area as the reference, we regulated all tiles into the same level of brightness. In addition, to get the location of immune cells' nuclei, we need to perform color deconvolution, an algorithm used to separate color space from immunohistochemical staining, on every tile of our datasets. To realize it, software ImageJ was used with a plugin called Color Deconvolution. In addition, in order to extract the Haralick texture of tumor cells, we used Positive Cell detection plugin in QuPath software to locate every tumor cell in each tile and use its batch process to get needed features.

*The Details of Feature Extraction:* In global color feature extraction, the region of interest (ROI) is a stained area. We recorded mean value, quantiles (25%, 50%, 70%) and higher order moments (variance, kurtosis, and skewness) in ROI of each channel in RGB and HSV as our global features. Besides, with GMM model, we perform image segmentation to each tile to divide the ROI into three clusters and record the corresponding features in every cluster as our local features. We located immune cells' nuclei after color deconvolution according to their size and grayscale and calculate the amount as the feature. As for the differentiation degree of tissue in tiles, we performed dilation, erosion and Hough Transformation, a method to identify outline similar to circle in images, to decide their differentiation degree. Since the more regular shapes exist, the more highly the tissue differentiates. Since we have recorded the tumor cell's location, we extract Haralick Features of each tumor cell in one tile and adopt the mean value of all cells' as the feature of this tile. Besides, we also recorded the count of tumor cell as our feature.

*The Random Forest Model for MSS and MSI Classification:* Our RF method was built and tested using Python version 3.7.1 with RandomForestClassifier in sklearn.ensemble library. [32] During training, 70% patients in every dataset were randomly selected and all of their tiles were used in training while the rest tiles were held out and used as test sets. There are some anomalous tiles in each dataset, i.e. blurred or color disorder, resulting in the loss of the information contained in them. Therefore, we disposed of all of them in every dataset. In addition, we also delete the tiles owning an extreme immune cell number (a value that significant in 1% level), since an extremely small number may represent the non-tumor area while a too large number represents lymphatic concentration area. In each forest, we set 500 trees in total and take Gini impurity as the criterion. The minimum number of samples to split an internal node is 2 and other parameters follow the default setting. In all cases, training and test sets were split on a patient level and no image tiles from test patients were present in any training sets. The visualization of the result in our model was realized with plotnine library in Python and AUCs were generated with sklearn.metrics library. And to further analyze the feature importance and interactions between features, we also used R version 3.5.1 with randomForest package[33] to rebuild that random forest and analyze and visualize the relations between different features with randomForestExplainer package[34].

Mean minimal depth is a measure that indicates the average depth of the feature's first occurrence in every tree. The importance ordering of features under it keeps highly consistent with the result from the permutation method. Pairwise interaction occurs when a split of one feature appears in maximal subtrees with respect to other features. To investigate the latent interaction between different features, we recorded all of such occurrences with one selected feature and calculated mean conditional minimal depth of variables concerning that feature. The times of occurrences and the decrease between the conditional minimal depth and corresponding mean minimal depth could be the indicators that reflect the degree of the interaction.

*Verify the Power of Features in Deep Learning:* We built the Resnet18 in Python 3.7 with TensorFlow-GPU 1.14.0 and Keras 2.3.0. In our network, patient-level AUCs are 0.77 in TCGA-CC-DX cohort, 0.81 in TCGA-CC-STAD cohort and 0.84 in TCGA-KR cohort. We eliminated the RGB mean differences between MSI and MSS groups in test set by adjusting the mean value in each tile in the test set to the mean value of all the tiles as a whole. The drops of AUCs after revaluation can verify the contribution of the RGB feature in the classification of the DL network.

**Supporting Information**
Supporting Information is available from the Wiley Online Library or from the author.

**Acknowledgements**
Wang's research is partially supported by NSFC (11771462), The National Key Research and Development Program of China (2018YFC1315400), The Key Research and Development Program of Guangdong, China (2019B020228001), and Pearl River S&T Nova Program of Guangzhou (201806010142). Liu’s research is supported by the National Natural Science Foundation of China (81902381). J.Z., W.W., R.L., and X.W. conceived and designed the study. J.Z., W.W., and Y.Z. performed the computational analysis. The paper was written by all co-authors. J.Z. and W.W. contributed equally to this work.

Received: ((will be filled in by the editorial staff))
Revised: ((will be filled in by the editorial staff))
Published online: ((will be filled in by the editorial staff))

References

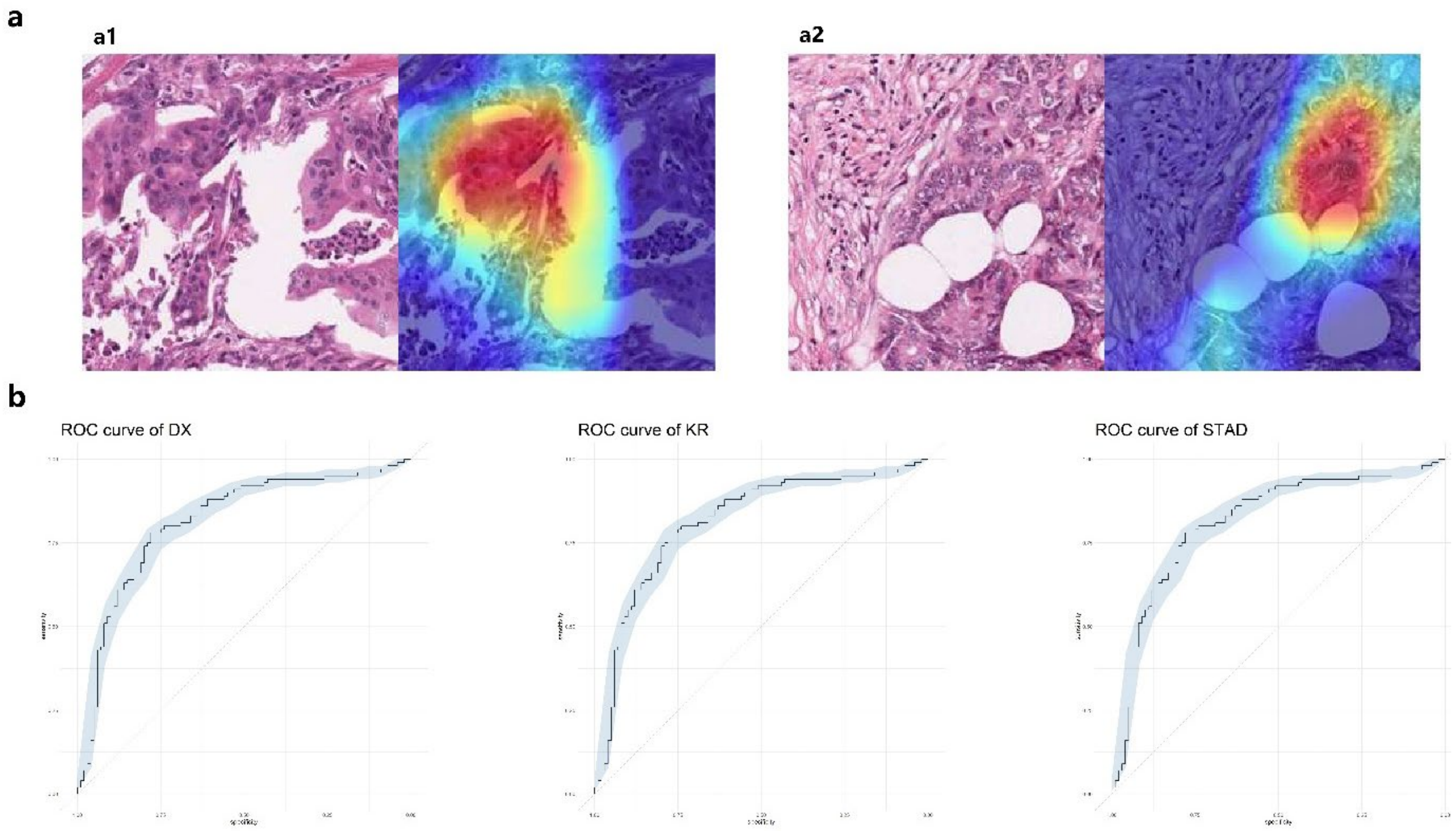

**Figure 1. (a) The original tile and the corresponding heatmap output by the GCAM. (a1)** and **(a2)** display tiles from the TCGA-CC-DX dataset labeled with MSI and MSS, respectively. In the heatmaps, the brighter region contributes more to the classification. For instance, the red one is the most highlighted area while the blue regions contribute limitedly. **(b) The workflow of studying pathological features in discriminating against MSI from MSS.** Five main steps, pretreatments, feature extraction, model training, patient-level predictions, and feature contributions analysis, were sequentially executed to improve the quality of image, generate pathological features, build statistical model, evaluate model performance, and measure features' contributions, respectively.

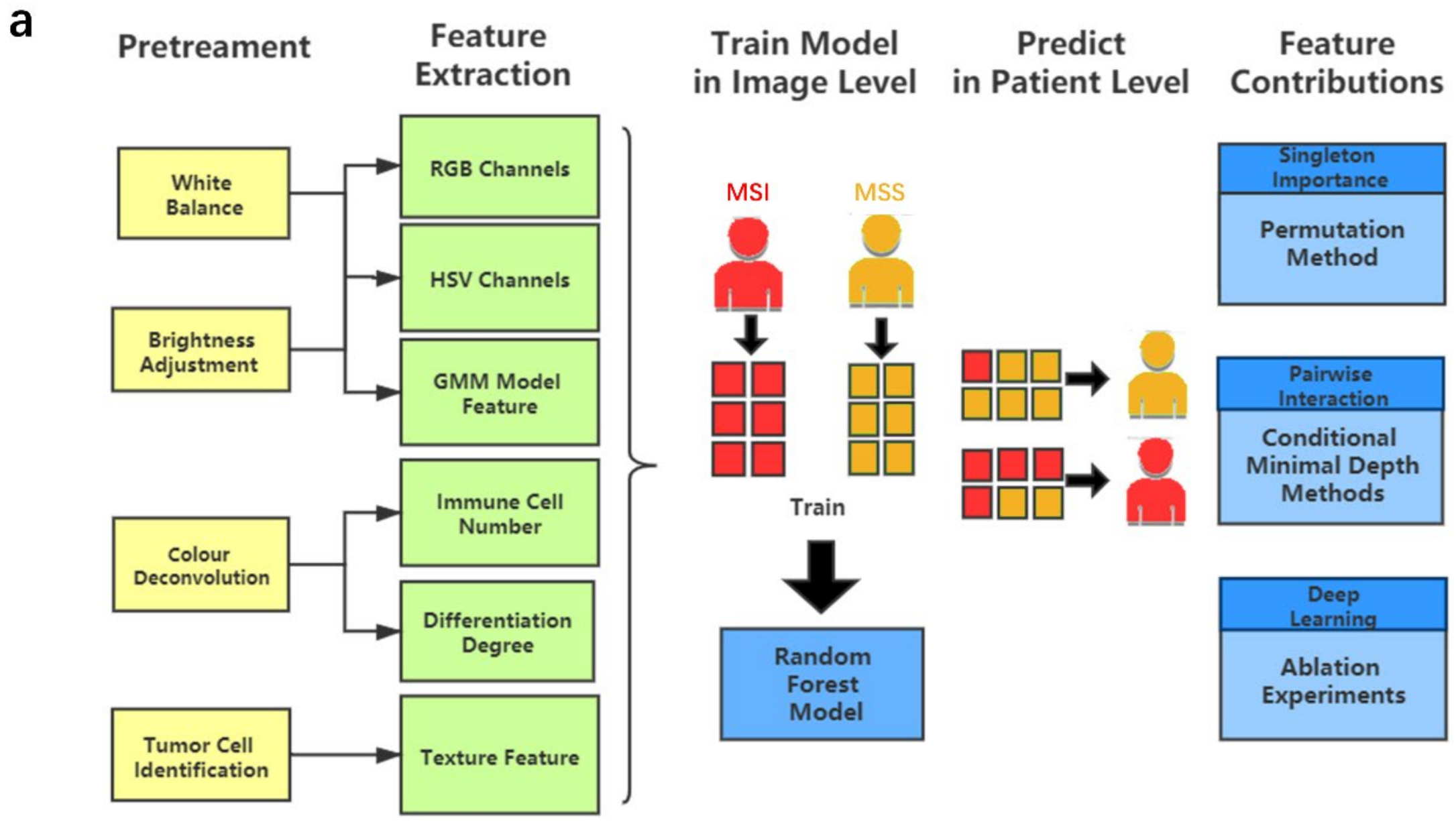

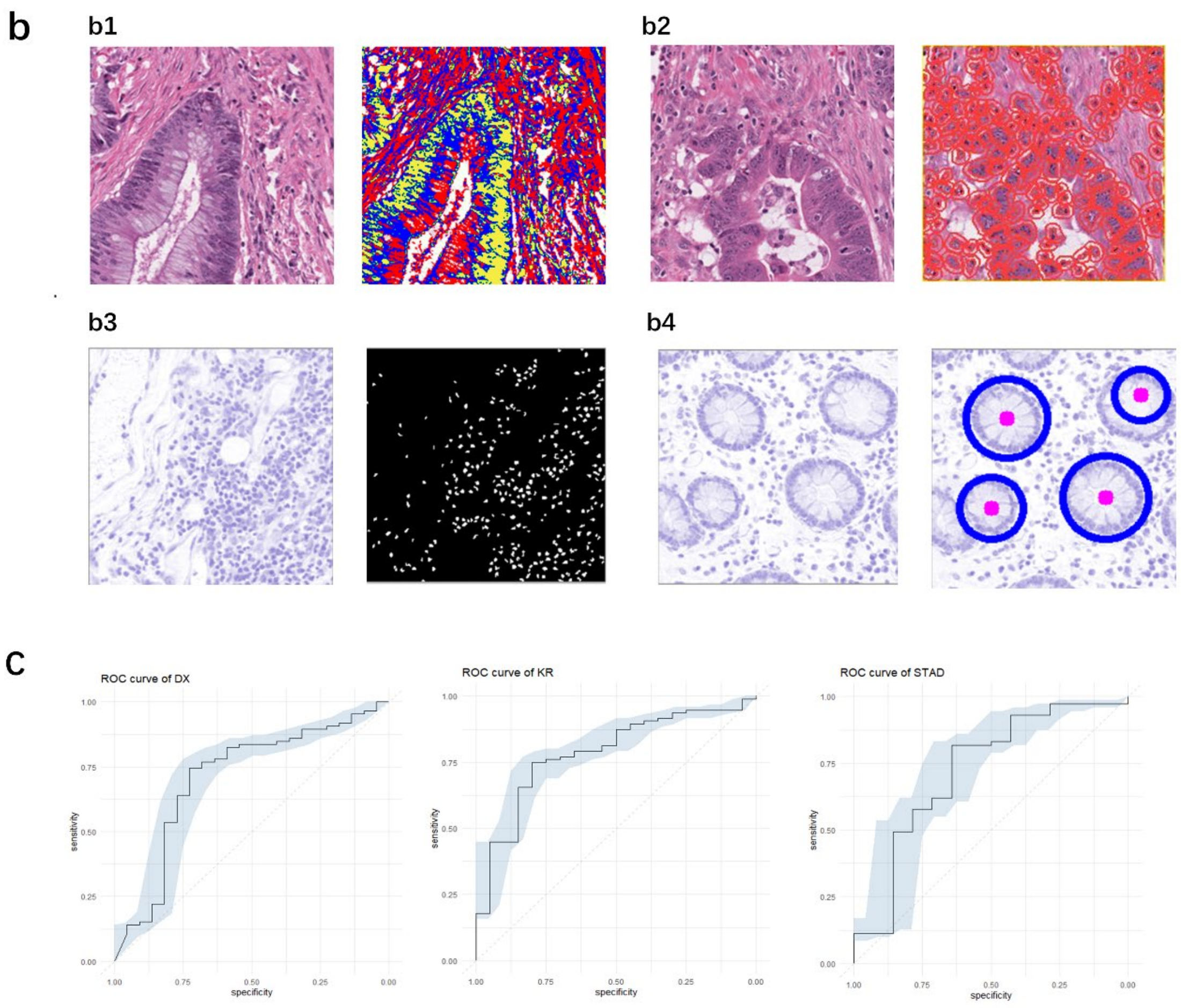

**Figure 2. (a) The workflow of studying pathological features in discriminating against MSI from MSS.** Five main steps, pretreatments, feature extraction, model training, patient-level predictions, and feature contributions analysis, were sequentially executed to improve the quality of image, generate pathological features, build statistical model, evaluate model performance, and measure features' contributions, respectively. **(b) Typical feature extraction result. (b1) Tumor cells detection before Haralick texture identification.** The left figure is an original tile while the right one is processed with tumor identification. Each red circle in the right tile indicates the boundary of one tumor cell. **(b2) GMM model for image segmentation.** The left figure is a tile from TCGA-CC-DX dataset, and its image-segmentation tiles processed by GMM method are shown in the right figure. The green part whose grayscale is lowest among three parts tends to be tumor tissue while the blue and red ones represent non-tumor tissue. **(b3) Infiltrating immune cells detection.** Detection of immune cells and we can calculate the connectivity domain. **(b4) The grading of differentiation.** Detect the circular similar arrangement in one slice and grade the degree of differentiation based on its amount. **(c) AUC values of RF models in three datasets.** Roc curves for classifying MSI versus MSS in each dataset.

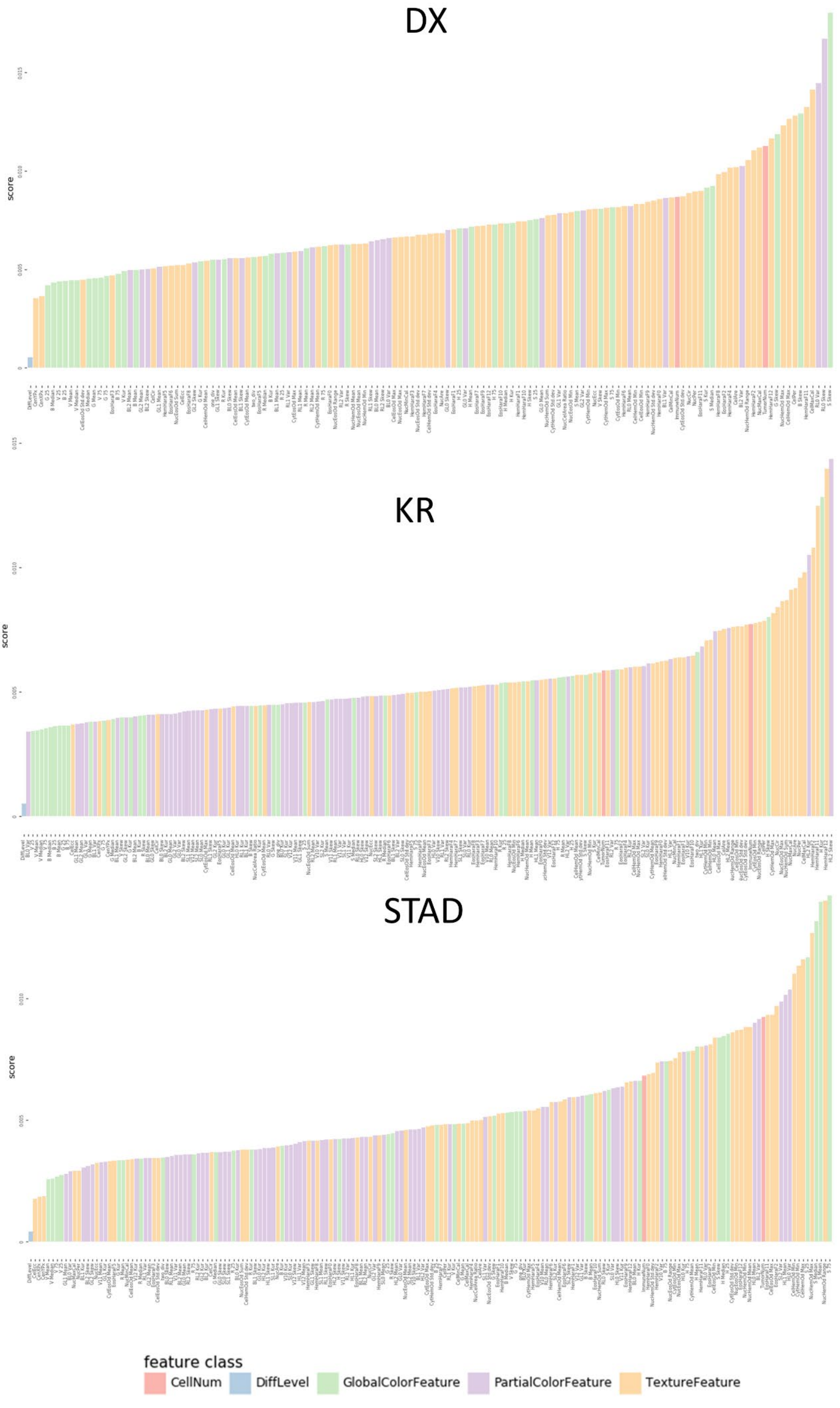

**Figure 3**. **(a) The permutation-based variable importance of each dataset.** The value of horizontal axis in each pic represents a degree of importance and the vertical axis displays corresponding variables' names. The rules of nomination are following the rules below: 1. Abbreviations of statistics: kur = kurtosis, skew = skewness, var = variance. 2. Combination of variable name: G-L2 var = the variance of G channel in the second level of the GMM model. 3. In texture features, every number follows F like 'F10' refers to one of feature from Haralick Texture. **(b) The distribution of minimal depth for the top ten variables in each dataset.** Minimal depth indicates the average depth one feature first occurs in every tree. The color bars from 0 to 10 indicates the depth that the feature first appears in each tree and the length of it indicates the count of such trees. NA means that the feature does not appear in that tree. We fill such missing value in mean depth calculation by setting the minimal depth of a variable in a tree that does not use it for splitting equally to the mean depth of trees.

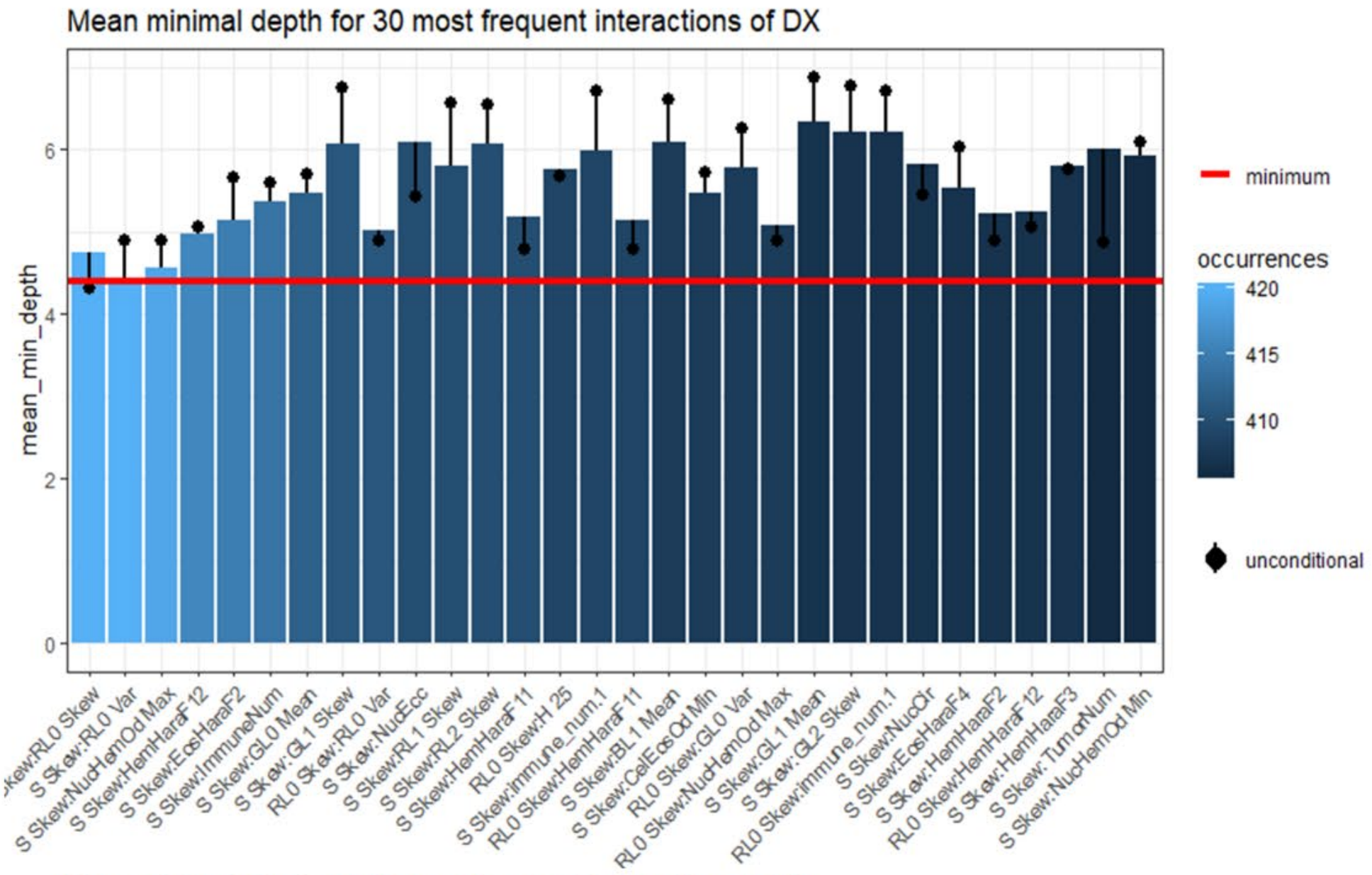
Mean minimal depth for 30 most frequent interactions of DX
mean_min_depth
minimum
occurrences
420
415
410
unconditional
0
2
4
6
S Skew:RL0 Var
S Skew:NucHemOd Max
S Skew:HemHaraF12
S Skew:EosHaraF2
S Skew:ImmuneNum
S Skew:GL0 Mean
S Skew:GL1 Skew
RL0 Skew:RL0 Var
S Skew:NucEcc
S Skew:RL1 Skew
S Skew:RL2 Skew
S Skew:HemHaraF11
RL0 Skew:H 25
S Skew:immune_num.1
RL0 Skew:HemHaraF11
S Skew:BL1 Mean
S Skew:CelEosOd Min
RL0 Skew:GL0 Var
RL0 Skew:NucHemOd Max
S Skew:GL1 Mean
S Skew:GL2 Skew
RL0 Skew:immune_num.1
S Skew:NucCir
S Skew:EosHaraF4
S Skew:HemHaraF2
RL0 Skew:HemHaraF12
S Skew:HemHaraF3
S Skew:TumorNum
S Skew:NucHemOd Min

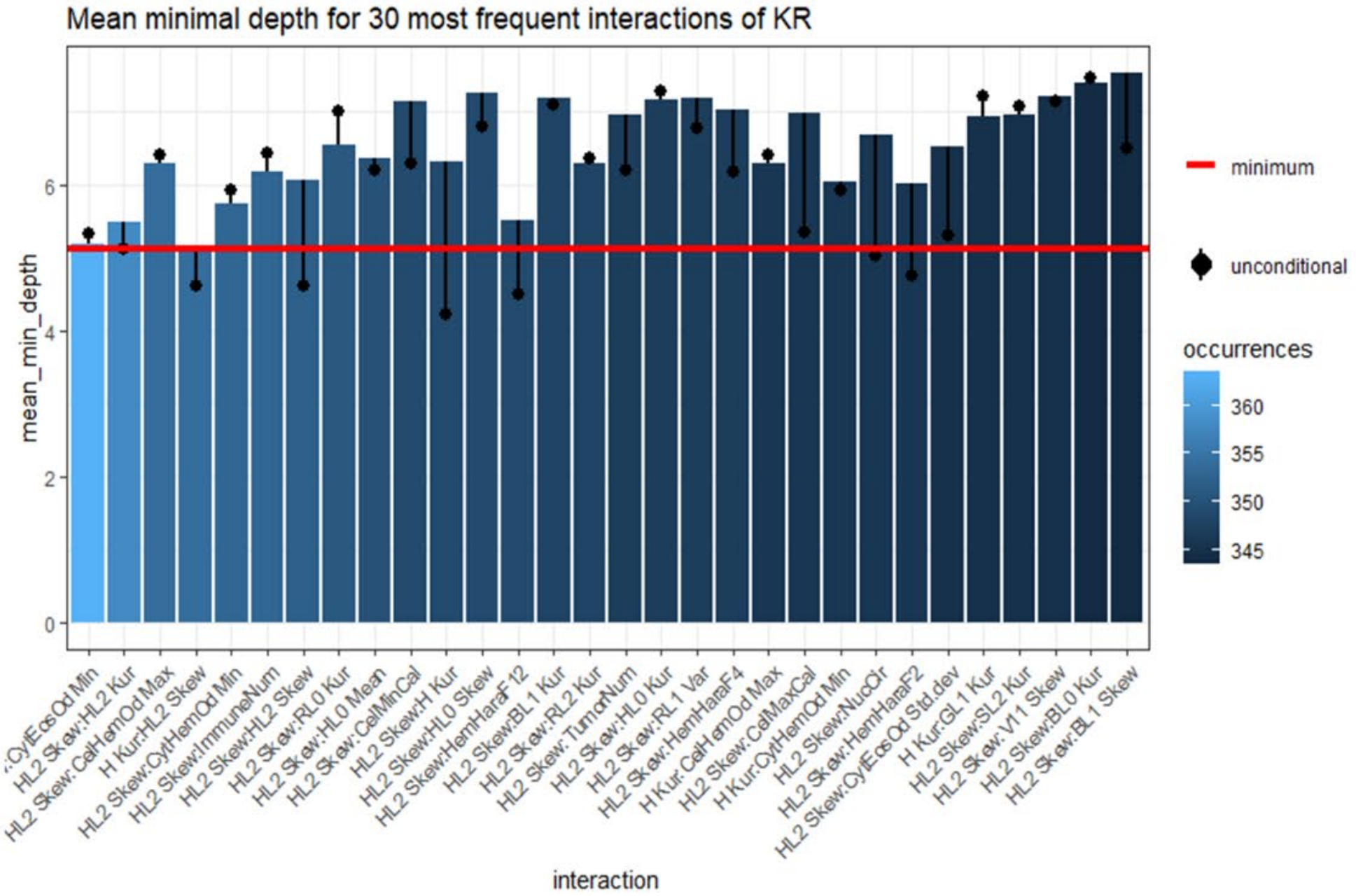
Mean minimal depth for 30 most frequent interactions of KR
mean_min_depth
minimum
unconditional
occurrences
360
355
350
345
0
2
4
6
interaction
HL2 Skew:HL2 Kur
HL2 Skew:CelHemOd Max
H Kur:HL2 Skew
HL2 Skew:CytHemOd Min
HL2 Skew:ImmuneNum
HL2 Skew:HL2 Skew
HL2 Skew:RL0 Kur
HL2 Skew:HL0 Mean
HL2 Skew:CelMinCal
HL2 Skew:H Kur
HL2 Skew:HL0 Skew
HL2 Skew:HemHaraF12
HL2 Skew:BL1 Kur
HL2 Skew:RL2 Kur
HL2 Skew:TumorNum
HL2 Skew:HL0 Kur
HL2 Skew:RL1 Var
HL2 Skew:HemHaraF4
H Kur:CelHemOd Max
HL2 Skew:CelMaxCal
H Kur:CytHemOd Min
HL2 Skew:NucCir
HL2 Skew:HemHaraF2
HL2 Skew:CytEosOd Std.dev
H Kur:GL1 Kur
HL2 Skew:SL2 Kur
HL2 Skew:VI1 Skew
HL2 Skew:BL0 Kur
HL2 Skew:BL1 Skew

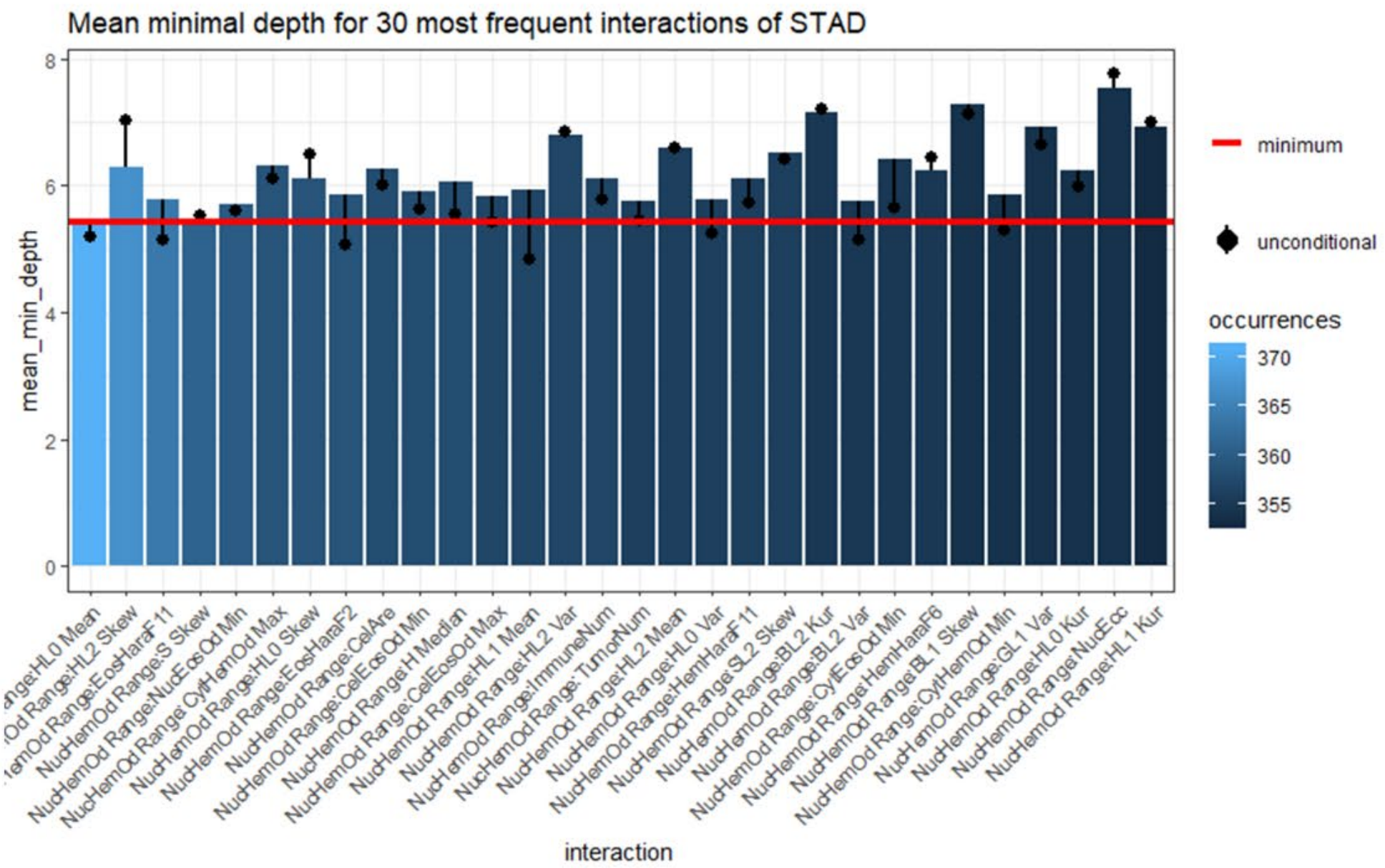
Mean minimal depth for 30 most frequent interactions of STAD
mean_min_depth
minimum
unconditional
occurrences
370
365
360
355
0
2
4
6
8
interaction

**Figure 4. The condition minimal depth of each datasets.** The color shade represents the occurrences of such interactions. The height of the bar indicates the conditional average depth of that feature and the height of the line dot means the unconditional average depth. The gap between them can reflect the degree of feature interaction.

An interpretable framework aims to assist doctors in MSI diagnosis. A deep learning model can not only make diagnosis for patients but also provide a heatmap to doctors such that they grasp the fundamental inner workings of model. A random forest model can help doctors with further research in feature interactions and enlighten them in diagnosis.

Jin Zhu, Wangwei Wu, Yuting Zhang, Shiyun Lin, Yukang Jiang, Ruixian Liu*, Xueqin Wang*

**Computational analysis of pathological image enables interpretable prediction for microsatellite instability**

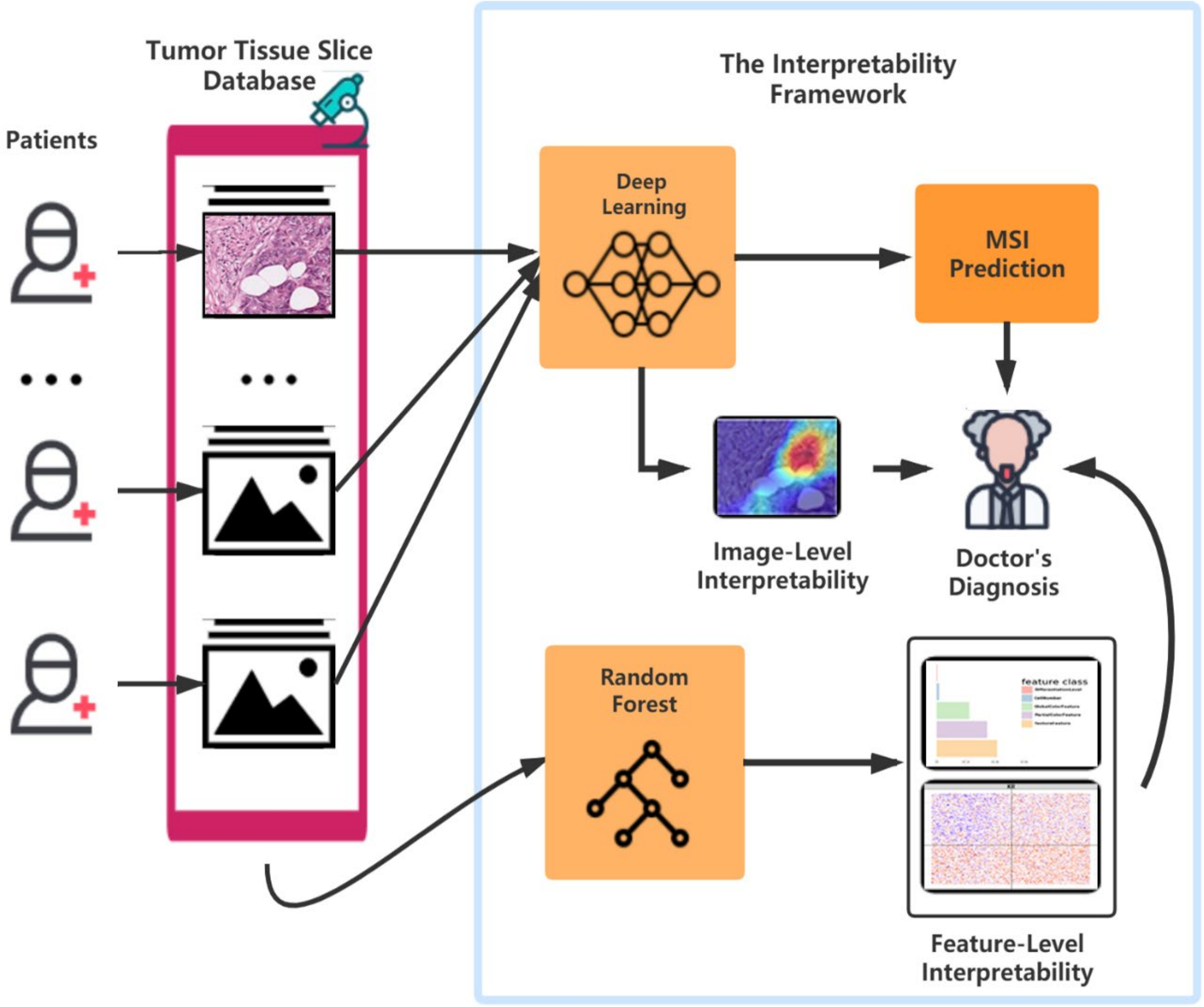

# Supporting Information

## Computational analysis of pathological image enables interpretable prediction for microsatellite instability

*Jin Zhu, Wangwei Wu, Yuting Zhang, Shiyun Lin, Yukang Jiang, Ruixian Liu*, Xueqin Wang**

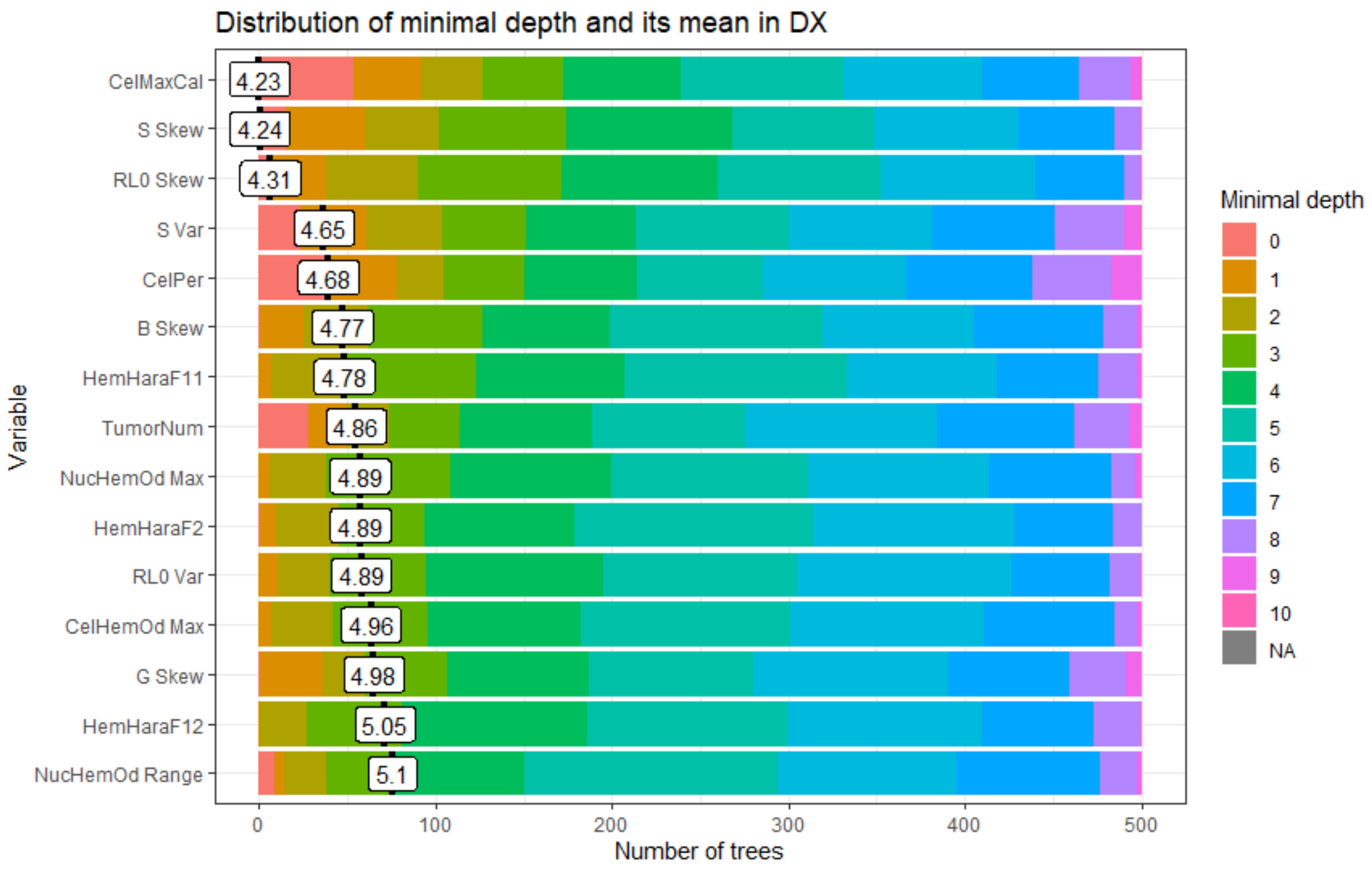

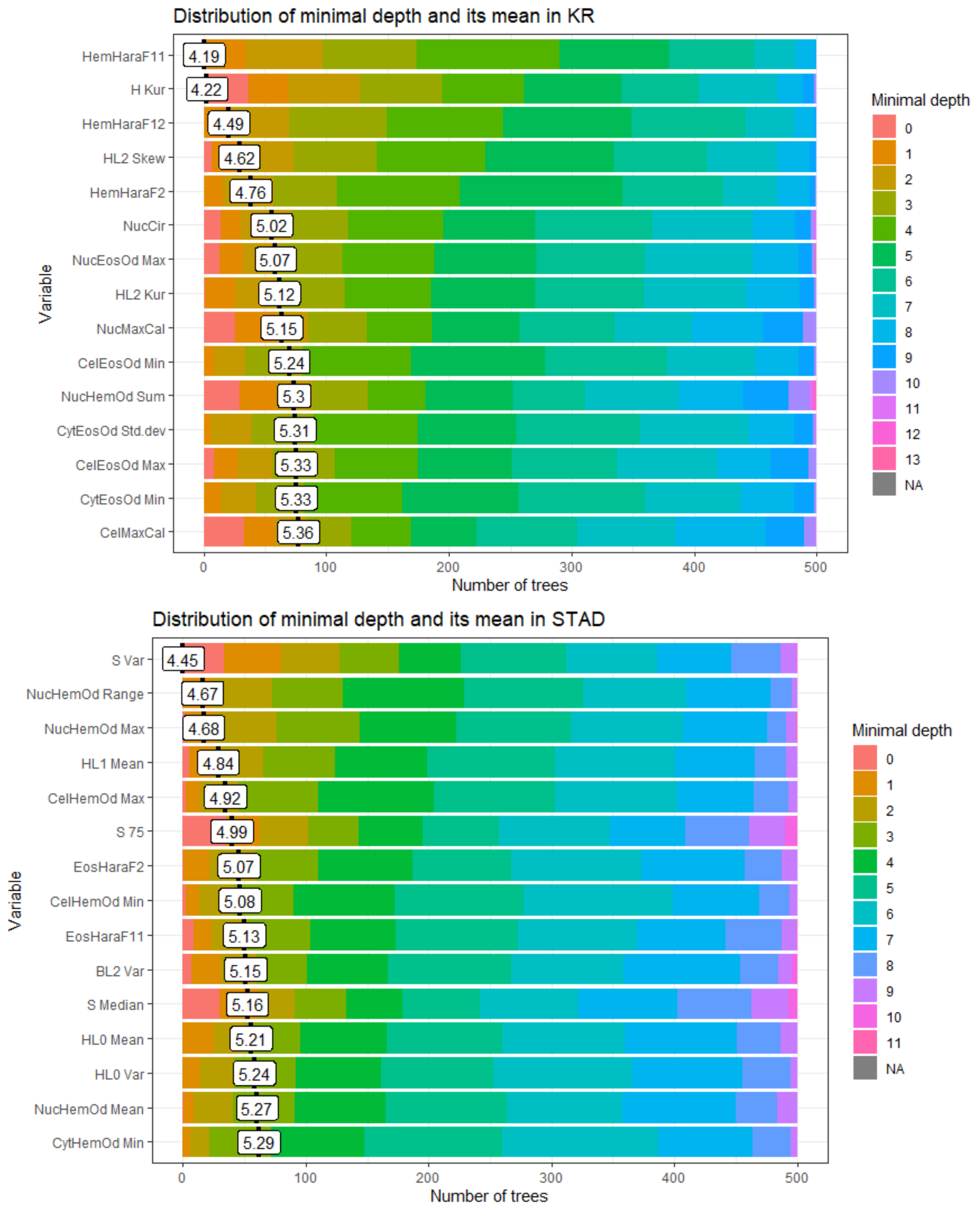

**Figure S1. The distribution of minimal depth for the top ten variables in each dataset.** Minimal depth indicates the average depth one feature first occurs in every tree. The color bars from 0 to 10 indicates the depth that the feature first appears in each tree and the length of it indicates the count of such trees. NA means that the feature does not appear in that tree. We fill such missing value in mean depth calculation by setting the minimal depth of a variable in a tree that does not use it for splitting equally to the mean depth of trees.

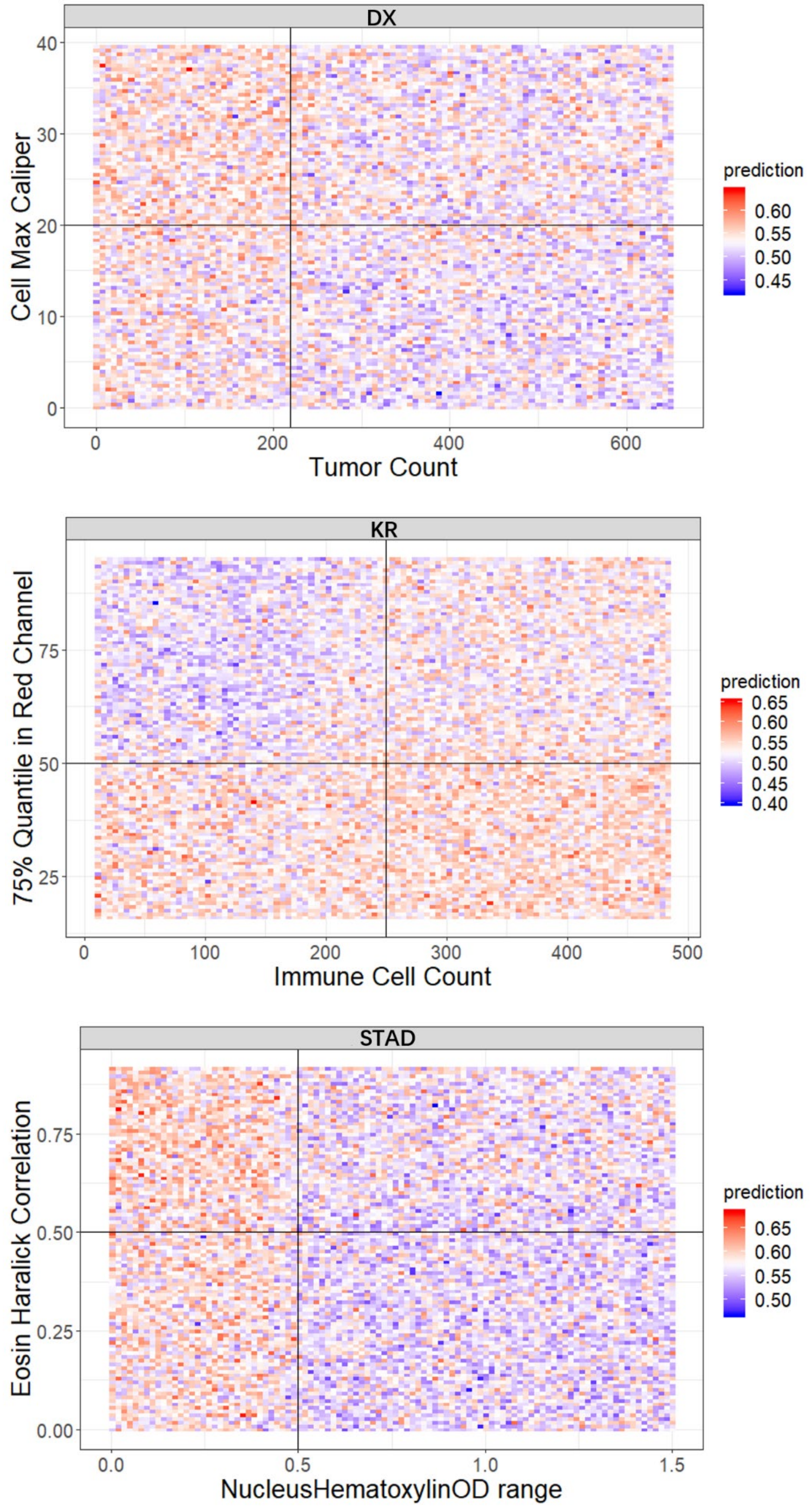

**Figure S2. The visualization of typical pairwise features' interaction in each dataset.** The forest predicts on a grid of values for the components of each interaction. The two axes represent the two corresponding features. The prediction value ranges from 0 to 1 with color

from blue to red. The bluer means a larger probability of MSI while the redder tends to be MSS.

**Table S3. The p-value of each feature in three dataset respectively under Wilcoxon Rank sum Test.** The nomination rules follow the ones in Fig3 of the main text. 2. The p-values are obtained from Wilcoxon Rank Sum Test, a nonparametric test focusing on hypothesize whether the samples X and Y come from the same population. 3. When the p-value are so small that exceeds the range of double-precision floating-point, it will turn out to be 0.

| Feature | P-Value DX | P-Value KR | P-Value STAD |
|---|---|---|---|
| r_var_l1 | 3.59E-267 | 1.00E-52 | 5.94E-34 |
| b_var_l2 | 9.80E-256 | 0 | 0 |
| r_var_l2 | 0.601107519 | 4.36E-278 | 0.157346181 |
| b_skew_l0 | 1.90E-54 | 2.56E-83 | 0 |
| g_var_l0 | 9.19E-126 | 2.06E-82 | 7.70E-76 |
| b_var_l1 | 0 | 1.34E-35 | 0 |
| b_skew_l2 | 0.1472704 | 4.63E-36 | 4.72E-13 |
| g_mean_l0 | 5.62E-104 | 0 | 4.15E-296 |
| r_skew_l2 | 7.14E-22 | 0 | 3.66E-136 |
| g_var_l1 | 8.58E-297 | 2.54E-170 | 0 |
| g_skew_l0 | 9.06E-127 | 2.52E-184 | 0 |
| g_mean_l1 | 0.350188506 | 0 | 1.57E-90 |
| r_mean_l1 | 5.01E-62 | 0 | 6.81E-56 |
| r_var_l0 | 8.71E-35 | 1.87E-307 | 2.49E-154 |
| g_skew_l1 | 0.353690612 | 4.80E-48 | 6.10E-39 |
| g_mean_l2 | 0.044152821 | 3.34E-05 | 0 |
| b_var_l0 | 1.95E-133 | 4.06E-115 | 2.80E-176 |
| r_skew_l1 | 5.71E-28 | 1.66E-125 | 0.035265802 |
| b_mean_l2 | 0.4905311 | 0.001619928 | 1.06E-292 |
| g_skew_l2 | 0.000266493 | 1.75E-28 | 2.79E-51 |
| g_var_l2 | 3.03E-179 | 0 | 1.82E-266 |
| b_skew_l1 | 2.85E-25 | 1.11E-23 | 1.17E-14 |
| b_mean_l1 | 0.000233127 | 0 | 4.59E-129 |
| r_skew_l0 | 5.21E-277 | 5.71E-29 | 5.39E-258 |
| b_mean_l0 | 4.69E-21 | 0 | 0 |
| r_mean_l2 | 5.77E-07 | 5.76E-16 | 0 |
| r_mean_l0 | 1.10E-217 | 0 | 0.261742479 |
| immune_num | 2.39E-252 | 2.43E-38 | 2.42E-17 |
| spot_num | 1.06E-25 | 1.74E-09 | 4.71E-149 |
| Nucleus_Eosin_OD_sum | 0 | 0 | 1.28E-237 |
| ROI_1_00_px_per_pixel_Eosin_Haralick_Contrast_F1_ | 1.45E-29 | 0 | 0 |
| ROI_1_00_px_per_pixel_Eosin_Haralick_Difference_entropy_F10_ | 5.08E-55 | 0 | 0 |
| Nucleus_Hematoxylin_OD_std_dev | 0.00E+00 | 0 | 1.21E-18 |
| Nucleus_Circularity | 0 | 0 | 0 |
| ROI_1_00_px_per_pixel_Hematoxylin_Haralick_Inverse_difference_moment_F4_ | 1.48E-11 | 1.40E-48 | 4.16E-12 |
| ROI_1_00_px_per_pixel_Hematoxylin_Haralick_Entropy_F8_ | 3.28E-186 | 0 | 0.181160256 |
| Nucleus_Area | 0 | 0 | 0 |
| Cell_Area | 0 | 0 | 3.27E-82 |
| ROI_1_00_px_per_pixel_Eosin_Haralick_Sum_entropy_F7_ | 5.97E-37 | 2.42E-223 | 0 |
| Cell_Hematoxylin_OD_std_dev | 2.15E-05 | 0.314876193 | 8.61E-125 |
| Nucleus_Eosin_OD_std_dev | 8.73E-11 | 0 | 0 |
| ROI_1_00_px_per_pixel_Hematoxylin_Haralick_Contrast_F1_ | 2.00E-134 | 0 | 3.53E-14 |
| Cell_Eosin_OD_std_dev | 1.35E-22 | 8.72E-68 | 0 |
| Nucleus_Eosin_OD_range | 3.25E-09 | 0 | 0 |
| ROI_1_00_px_per_pixel_Eosin_Haralick_Entropy_F8_ | 0.431883476 | 9.67E-225 | 0 |
| Cytoplasm_Eosin_OD_std_dev | 5.80E-108 | 2s.64E-78 | 1.96E-07 |
| ROI_1_00_px_per_pixel_Eosin_Haralick_Angular_second_moment_F0_ | 7.48E-07 | 2.45E-132 | 0 |
| ROI_1_00_px_per_pixel_Hematoxylin_Haralick_Difference_variance_F9_ | 0.000164701 | 1.25E-20 | 3.13E-05 |
| Cell_Hematoxylin_OD_min | 1.59E-59 | 4.45E-103 | 0.200857364 |
| ROI_1_00_px_per_pixel_Hematoxylin_Haralick_Sum_of_squares_F3_ | 3.15E-28 | 7.08E-111 | 2.74E-213 |
| Cell_Max_caliper | 0 | 0 | 2.07E-76 |
| Nucleus_Hematoxylin_OD_mean | 1.71E-93 | 0 | 3.39E-80 |
| Nucleus_Max_caliper | 0 | 0 | 6.02E-297 |
| Nucleus_Hematoxylin_OD_range | 1.90E-209 | 0 | 0 |
| Cell_Eccentricity | 0 | 0 | 7.21E-51 |
| ROI_1_00_px_per_pixel_Hematoxylin_Haralick_Angular_second_moment_F0_ | 2.87E-45 | 2.55E-207 | 2.78E-14 |
| Nucleus_Eosin_OD_mean | 9.89E-19 | 2.31E-22 | 3.81E-06 |
| ROI_1_00_px_per_pixel_Eosin_Haralick_Information_measure_of_correlation_2_F12_ | 1.01E-07 | 1.11E-110 | 0 |
| Cell_Hematoxylin_OD_mean | 1.56E-189 | 0 | 0.000401154 |
| Cell_Hematoxylin_OD_max | 5.01E-97 | 8.62E-74 | 0 |
| Nucleus_Perimeter | 0 | 0 | 0 |
| Cytoplasm_Hematoxylin_OD_std_dev | 0 | 0 | 7.47E-26 |
| Cell_Eosin_OD_mean | 3.01E-08 | 1.24E-89 | 0 |
| Cell_Eosin_OD_min | 4.55E-18 | 1.54E-09 | 1.86E-304 |
| Cytoplasm_Eosin_OD_mean | 1.86E-14 | 1.46E-100 | 0 |
| ROI_1_00_px_per_pixel_Hematoxylin_Haralick_Information_measure_of_correlation_1_F11_ | 9.75E-137 | 1.02E-38 | 3.85E-10 |
| ROI_1_00_px_per_pixel_Hematoxylin_Haralick_Information_measure_of_correlation_2_F12_ | 6.32E-90 | 3.58E-23 | 3.05E-09 |
| ROI_1_00_px_per_pixel_Eosin_Haralick_Correlation_F2_ | 2.71E-174 | 2.01E-23 | 1.89E-273 |
| Centroid_Y_px | 0.020626705 | 0.645007505 | 0.441953604 |
| Centroid_X_px | 0.472150747 | 0.315233167 | 0.602055352 |
| Nucleus_Min_caliper | 0 | 0 | 0 |
| Cytoplasm_Hematoxylin_OD_min | 3.15E-59 | 7.09E-102 | 0.034861466 |
| ROI_1_00_px_per_pixel_Eosin_Haralick_Sum_of_squares_F3_ | 1.44E-10 | 1.68E-83 | 0 |
| Nucleus_Eosin_OD_min | 1.66E-07 | 7.85E-104 | 0 |
| Nucleus_Hematoxylin_OD_sum | 0 | 0 | 9.37E-48 |
| ROI_1_00_px_per_pixel_Hematoxylin_Haralick_Sum_entropy_F7_ | 2.07E-163 | 0 | 2.85E-08 |
| Nucleus_Cell_area_ratio | 3.41E-287 | 0 | 0 |
| ROI_1_00_px_per_pixel_Eosin_Haralick_Information_measure_of_correlation_1_F11_ | 2.32E-28 | 3.86E-11 | 0 |
| Cell_Circularity | 0 | 0 | 0 |
| Cytoplasm_Eosin_OD_min | 2.08E-17 | 2.87E-08 | 0 |
| ROI_1_00_px_per_pixel_Hematoxylin_Haralick_Correlation_F2_ | 0.511355632 | 2.90E-158 | 4.57E-165 |
| ROI_1_00_px_per_pixel_Hematoxylin_Haralick_Difference_entropy_F10_ | 2.72E-44 | 4.38E-19 | 0.511991852 |
| Nucleus_Eosin_OD_max | 0.17439814 | 0 | 0 |
| Cytoplasm_Hematoxylin_OD_mean | 7.45E-160 | 0 | 4.34E-12 |
| count | 0 | 0 | 8.92E-05 |
| Nucleus_Hematoxylin_OD_min | 1.42E-67 | 0 | 1.45E-16 |
| ROI_1_00_px_per_pixel_Eosin_Haralick_Inverse_difference_moment_F4_ | 1.44E-26 | 0 | 0 |
| Cell_Eosin_OD_max | 0.001632192 | 0 | 0 |
| Nucleus_Hematoxylin_OD_max | 4.29E-108 | 3.85E-76 | 0 |
| ROI_1_00_px_per_pixel_Hematoxylin_Haralick_Sum_average_F5_ | 2.35E-176 | 0 | 5.99E-05 |
| ROI_1_00_px_per_pixel_Eosin_Haralick_Sum_variance_F6_ | 0.005079665 | 2.15E-187 | 0 |
| Cytoplasm_Hematoxylin_OD_max | 0 | 0 | 6.21E-80 |
| Nucleus_Eccentricity | 0 | 2.80E-282 | 5.84E-12 |
| ROI_1_00_px_per_pixel_Eosin_Haralick_Sum_average_F5_ | 5.64E-08 | 3.80E-76 | 0 |
| Cell_Perimeter | 0 | 0 | 1.29E-79 |
| Cell_Min_caliper | 0 | 0 | 5.78E-55 |
| ROI_1_00_px_per_pixel_Hematoxylin_Haralick_Sum_variance_F6_ | 6.67E-05 | 7.86E-47 | 5.69E-274 |
| Cytoplasm_Eosin_OD_max | 0.59124779 | 1.63E-58 | 9.46E-269 |
| ROI_1_00_px_per_pixel_Eosin_Haralick_Difference_variance_F9_ | 2.89E-43 | 0 | 0 |
| h_range | 3.93E-10 | 3.60E-289 | 2.66E-13 |
| r_median | 0.021810527 | 0 | 3.52E-231 |
| r_kur | 2.07E-55 | 0 | 6.49E-53 |
| s_skew | 0 | 0 | 0 |
| r_75 | 0.569351629 | 0 | 3.57E-223 |
| g_25 | 0.006391458 | 0 | 5.13E-164 |
| s_median | 1.51E-120 | 8.88E-293 | 0 |
| b_range | 2.42E-280 | 0 | 0 |
| b_75 | 2.77E-66 | 0 | 0 |
| s_kur | 5.57E-09 | 0.0040273 | 0.01395365 |
| g_var | 2.86E-29 | 0 | 0 |
| v_mean | 4.57E-07 | 0 | 1.05E-106 |
| r_25 | 6.19E-36 | 0 | 8.25E-87 |
| g_skew | 1.47E-179 | 0 | 0 |
| h_median | 2.80E-249 | 0 | 6.97E-236 |
| h_25 | 0 | 0 | 8.75E-281 |
| h_mean | 4.53E-283 | 0 | 5.75E-198 |
| v_25 | 5.37E-19 | 0 | 2.25E-115 |
| v_75 | 0.696136204 | 0 | 3.46E-261 |
| v_range | 7.50E-241 | 2.13E-78 | 7.63E-282 |
| r_var | 7.72E-53 | 2.35E-66 | 0 |
| b_median | 1.53E-62 | 0 | 0 |
| h_skew | 1.87E-273 | 0 | 6.63E-105 |
| s_25 | 3.92E-112 | 0 | 0 |
| r_skew | 5.99E-17 | 0 | 0 |
| g_range | 6.53E-85 | 0 | 0 |
| v_kur | 0.431313439 | 0 | 0 |
| r_range | 0.060903193 | 1.88E-44 | 5.18E-143 |
| h_kur | 2.19E-212 | 0 | 2.57E-60 |
| g_mean | 3.16E-14 | 0 | 0 |
| r_mean | 1.16E-13 | 0 | 1.67E-95 |
| v_var | 6.21E-20 | 1.57E-127 | 0 |
| b_var | 1.00E-80 | 6.25E-178 | 0 |
| b_25 | 0.000152581 | 0 | 3.25E-200 |
| v_median | 0.934911866 | 0 | 3.06E-266 |
| s_75 | 5.95E-43 | 1.37E-103 | 1.60E-251 |
| b_mean | 4.68E-24 | 0 | 0 |
| s_range | 5.17E-224 | 0 | 0 |
| g_median | 5.29E-43 | 0 | 0 |
| b_kur | 4.25E-08 | 6.93E-162 | 2.86E-150 |
| s_mean | 6.72E-65 | 2.22E-238 | 0 |
| s_var | 2.11E-80 | 3.27E-205 | 0 |
| g_75 | 3.52E-37 | 0 | 0 |
| v_skew | 2.26E-07 | 0 | 0 |
| b_skew | 9.97E-167 | 0 | 0 |
| g_kur | 1.53E-60 | 5.25E-35 | 2.60E-69 |
| h_75 | 1.07E-182 | 0 | 1.97E-125 |
| h_var | 1.12E-24 | 0.027584937 | 5.98E-09 |

[1] R. J. Hause, C. C. Pritchard, J. Shendure, S. J. Salipante, *Nature medicine*, **2016**. *22*, 1342.
[2] S. Popat, R. Hubner, R. S. Houlston, *Journal of Clinical Oncology*, **2005**. *23*, 609.
[3] X. Sagaert, E. V. Cutsem, S. Tejpar, H. Prenen, G. D. Hertogh, *Journal of Clinical Oncology*, **2014**. *32*, 495.
[4] T. Y. Wong, N. M. Bressler, *The Journal of the American Medical Association*, **2016**. *316*, 2366.
[5] A. Serag, A. Ion-Margineanu, H. Qureshi, R. McMillan, M.-J. Saint Martin, J. Diamond, P. O'Reilly, P. Hamilton, *Frontiers in Medicine*, **2019**. *6*.
[6] O. Iizuka, F. Kanavati, K. Kato, M. Rambeau, K. Arihiro, M. Tsuneki, *Scientific Reports*, **2020**. *10*, 1.
[7] Y. Bar, I. Diamant, L. Wolf, H. Greenspan, presented at *Medical Imaging 2015: Computer-Aided Diagnosis*, **2015**.
[8] J. N. Kather, A. T. Pearson, N. Halama, D. Jäger, J. Krause, S. H. Loosen, A. Marx, P. Boor, F. Tacke, U. P. Neumann, *Nature medicine*, **2019**. *25*, 1054.
[9] W. J. Murdoch, C. Singh, K. Kumbier, R. Abbasi-Asl, B. Yu, *Proceedings of the National Academy of Sciences*, **2019**. *116*, 22071.
[10] N. C. Jacobson, K. H. Bentley, A. Walton, S. B. Wang, R. G. Fortgang, A. J. Millner, G. Coombs III, A. M. Rodman, D. D. Coppersmith, *Bulletin of the World Health Organization*, **2020**. *98*.
[11] E. Vayena, A. Blasimme, I. G. Cohen, *PLoS medicine*, **2018**. *15*.
[12] A. J. Schaumberg, W. C. Juarez-Nicanor, S. J. Choudhury, L. G. Pastrián, B. S. Pritt, M. P. Pozuelo, R. S. Sánchez, K. Ho, N. Zahra, B. D. Sener, *Modern Pathology*, **2020**.
[13] A. Vellido, *Kidney Diseases*, **2019**. *5*, 11.
[14] S. L. Piano, *Humanities Social Sciences Communications*, **2020**. *7*, 1.
[15] R. Elshawi, M. H. Al-Mallah, S. Sakr, *BMC medical informatics decision making*, **2019**. *19*.
[16] E. Lee, J.-S. Choi, M. Kim, H.-I. Suk, A. s. D. N. Initiative, *NeuroImage*, **2019**. *202*.
[17] C. G. A. Network, *Nature*, **2012**. *487*, 330.
[18] C. G. A. R. Network, *Nature*, **2014**. *513*, 202.
[19] K. He, X. Zhang, S. Ren, J. Sun, presented at *Proceedings of the IEEE conference on computer vision and pattern recognition*, **2016**.
[20] R. R. Selvaraju, M. Cogswell, A. Das, R. Vedantam, D. Parikh, D. Batra, presented at *Proceedings of the IEEE international conference on computer vision*, **2017**.
[21] A. C. Ruifrok, D. A. Johnston, *Analytical quantitative cytology histology*, **2001**. *23*, 291.
[22] G. J. McLachlan, D. Peel, *Finite mixture models*. John Wiley & Sons **2004**.
[23] H. Yuen, J. Princen, J. Illingworth, J. Kittler, *Image vision computing*, **1990**. *8*, 71.
[24] P. Bankhead, M. B. Loughrey, J. A. Fernández, Y. Dombrowski, D. G. McArt, P. D. Dunne, S. McQuaid, R. T. Gray, L. J. Murray, H. G. Coleman, *Scientific reports*, **2017**. *7*, 1.
[25] L. Breiman, *Machine learning*, **2001**. *45*, 5.
[26] G. Biau, E. Scornet, *Test*, **2016**. *25*.
[27] H. Ishwaran, U. B. Kogalur, E. Z. Gorodeski, A. J. Minn, M. S. Lauer, *Journal of the American Statistical Association*, **2010**. *105*, 205.

[28] D. Denisko, M. M. Hoffman, *Proceedings of the National Academy of Sciences*, **2018**. *115*, 1690.
[29] J. K. Greenson, S.-C. Huang, C. Herron, V. Moreno, J. D. Bonner, L. P. Tomsho, O. Ben-Izhak, H. I. Cohen, P. Trougouboff, J. Bejhar, *The American journal of surgical pathology*, **2009**. *33*, 126.
[30] L. Nguyen, A. B. Tosun, J. L. Fine, A. V. Lee, D. L. Taylor, S. C. Chennubhotla, *IEEE transactions on medical imaging*, **2017**. *36*, 1522.
[31] L. R. Bonetti, V. Barresi, S. Bettelli, F. Domati, C. Palmiere, *Diagnostic pathology*, **2016**. *11*, 31.
[32] F. Pedregosa, G. Varoquaux, A. Gramfort, V. Michel, B. Thirion, O. Grisel, M. Blondel, P. Prettenhofer, R. Weiss, V. Dubourg, *Journal of machine Learning research*, **2011**. *12*, 2825.
[33] A. L. a. M. Wiener, *R News*, **2002**. *2*, 18.
[34] A. Paluszynska, P. Biecek *randomForestExplainer: Explaining and visualizing random forests in terms of variable importance* in *R package version 0.9*. **2017**.